\documentclass[11pt]{article}

\usepackage[final]{acl}
\usepackage{amsmath}
\usepackage{times}
\usepackage{latexsym}

\usepackage[T1]{fontenc}

\usepackage[utf8]{inputenc}

\usepackage{microtype}

\usepackage{inconsolata}
\usepackage{amsfonts}
\usepackage[most]{tcolorbox}

\usepackage{graphicx}
\usepackage{subcaption}
\usepackage{booktabs}
\usepackage{multirow}
\usepackage[table]{xcolor}
\usepackage{makecell}
\usepackage{hyperref}
\title{Reasoning Error from Known Fact: \\Step-Level Self-Consistency Group Relative Policy Optimization for LLMs}

\author{
\textbf{Xiaomeng Hu,
~Jiaqi Hu,
~Hao Chen,
~Qi Zhang,} \\
\textbf{~Zhanming Shen,
~Wentao Ye,
~Junbo Zhao} \\
{Zhejiang University}\\
\texttt{\{xm.hu,j.zhao\}@zju.edu.cn}\\
}

\begin{document}
\maketitle
\begin{abstract}
With the rapid advancement of large language models (LLMs), modern systems not only possess strong foundational capabilities and extensive knowledge, but can also solve complex problems via long, multi-step reasoning. However, as reasoning traces become longer, LLMs may produce a substantial amount of hallucinated content during the reasoning process, which is often difficult to detect. In this work, we conduct a fine-grained analysis of hallucinations arising in LLM reasoning and find that the reasoning traces are particularly prone to \emph{Context-Sensitive Factual Hallucinations}—cases where the model actually has the relevant knowledge, yet makes factual errors due to contextual interference during reasoning. To address this issue, we propose \textbf{S}tep-level \textbf{S}elf-\textbf{C}onsistency \textbf{G}roup \textbf{R}elative \textbf{P}olicy \textbf{O}ptimization (SSC-GRPO), which assigns step-level rewards to reasoning traces by computing self-consistency scores of individual steps across multiple rollouts. Compared with prior methods, SSC-GRPO achieves state-of-the-art performance on both mathematical reasoning benchmarks and hallucination leaderboards. Our results offer a new perspective for detecting and mitigating hallucinations in the reasoning process of large language models.
\end{abstract}

\section{Introduction}

In recent years, with the continuous advancement of large language models (LLMs), remarkable performance has been achieved across many domains~\citep{achiam2023gpt,team2024gemini,grattafiori2024llama}. Existing general-purpose models possess strong foundational capabilities, and their domain knowledge is increasingly comprehensive. Moreover, to better solve complex problems, reasoning models have undergone rapid development~\citep{yang2025qwen3, guo2025deepseek}. The \emph{test-time scaling} principle suggests that models can analyze questions step by step via long chains of thought and ultimately derive final answers~\citep{wu2024inference}.

However, LLMs suffer from the \emph{hallucination} problem, in which models generate confident but incorrect statements~\citep{ji2023survey}. Since hallucinations severely hinder the development and deployment of LLMs, the detection and mitigation of hallucinations have become central research topics in recent years~\citep{zhang2025siren}. As we move beyond general tasks and consider reasoning models, hallucinations occurring during the reasoning process present fundamentally different challenges~\citep{li2024fg}. The long reasoning chains and complex contextual dependencies render traditional hallucination detection methods ineffective~\citep{cheng2025chain}, necessitating a deeper study of hallucinations within the reasoning process.

\begin{figure*}[th]
    \centering
  \includegraphics[width=0.92\linewidth]{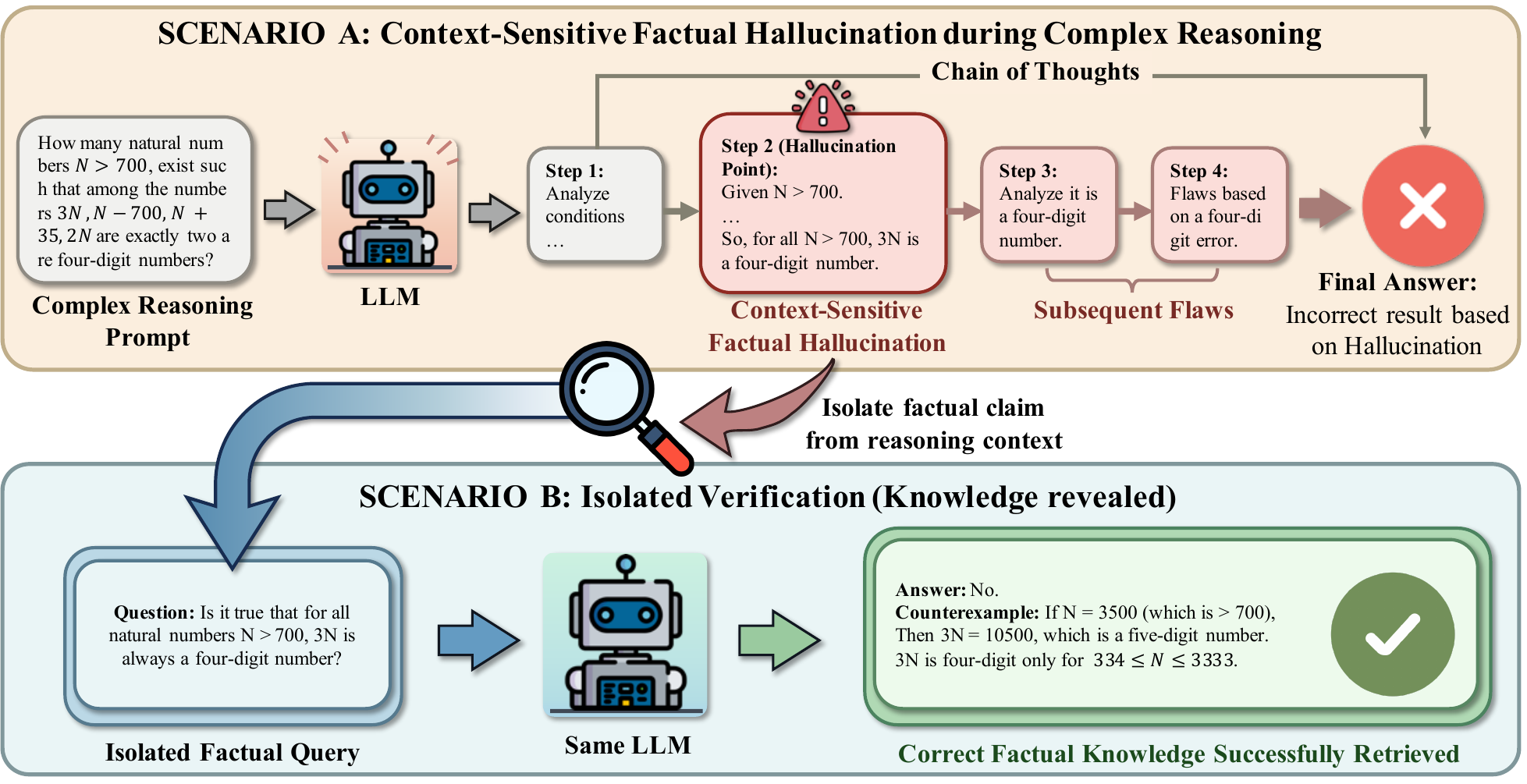}
  \caption {\textbf{Context-Sensitive Factual Hallucination in Large Language Models.}
During the reasoning process, large language models may produce factual errors that subsequently lead to incorrect final answers, as illustrated in Scenario A. However, when the erroneous part is extracted and reformulated as an independent question, the model is able to answer it correctly, as shown in Scenario B. This phenomenon indicates that the model actually possesses the relevant factual knowledge, but generates hallucinations due to the influence of the reasoning context.}
  \label{fig:context-hallu}
  \vspace{-10pt}
\end{figure*}

To better study hallucinations during reasoning, we conduct a fine-grained categorization of observed hallucination types. Through detailed case analyses, we find that current models frequently exhibit \textbf{\emph{context-sensitive factual hallucinations}}: factual errors that arise within the reasoning process even though the model actually knows the relevant knowledge, but is misled by the surrounding reasoning context, as shown in  Figure~\ref{fig:context-hallu}. Specifically, if we extract the hallucinated segment and rewrite it as an independent question to query the same model, it can answer correctly; yet within the original reasoning context, it produces a factual mistake. By analyzing hallucination categories during reasoning, we observe that context-sensitive hallucinations account for the overwhelming majority, as shown in Figure~\ref{fig:hallu_analysis}.

To address this issue, we propose \textbf{S}tep-Level \textbf{S}elf-\textbf{C}onsistency \textbf{G}roup \textbf{R}elative \textbf{P}olicy \textbf{O}ptimization(SSC-GRPO), a reinforcement-learning-based method to mitigate hallucinations in the model's reasoning process. In the standard GRPO algorithm, the model performs multiple rollouts and samples a set of outputs; similarly, self-consistency-based hallucination detection also relies on multiple samples. \textbf{This indicates that GRPO naturally aligns with self-consistency hallucination detection}. Furthermore, because the hallucinated content during reasoning is knowledge the model actually possesses, self-consistency detection can locate where hallucinations occur \emph{without external knowledge}, enabling step-level rewards for GRPO. Following the idea of \textsc{SelfCheckGPT}~\citep{manakul2023selfcheckgpt}, we prompt the model being trained to judge its own consistency and obtain a consistency score, which is used as the reward for each reasoning step. We then use step-level consistency rewards to construct clipped multiplicative weights for the rollout-level advantages. Finally, SSC-GRPO optimizes the model using the resulting step-level advantages.

To validate the effectiveness of our method, we conduct detailed experiments on both hallucination and mathematics tasks. We train models with SSC-GRPO on Qwen3-4B-Base, Qwen3-4B-Instruct and Llama3-8B-Instruct, and compare against a GRPO baseline and other methods. The results show that our approach achieves state-of-the-art performance on both math and hallucination tasks, outperforming the baseline by 1.8\%. Furthermore, we analyze consistency dynamics during training and find that consistency steadily increases, while the number of inconsistent steps decreases, providing additional evidence for the effectiveness of our approach.

In summary, our contributions are as follows:
\begin{itemize}
  \item We identify \textbf{context-sensitive factual hallucinations} in the reasoning process of existing models: even when the model knows the relevant facts, it can be influenced by context and make factual errors during reasoning. We further demonstrate experimentally that this type of hallucination dominates in the reasoning process.
  \item We propose \textbf{SSC-GRPO}, which uses self-consistency to provide \textbf{step-level rewards} for reasoning steps, thereby mitigating context-sensitive factual hallucinations.
  \item We design extensive experiments on both mathematics and hallucination tasks to verify the effectiveness of our method.
\end{itemize}

\section{Related Work}

\subsection{Hallucination in Large Language Models }
As large language models (LLMs) evolve rapidly, hallucination—the generation of fluent yet factually unfounded content—has become a primary concern, severely impeding the reliability and practical deployment of these systems~\citep{zhang2025siren}.  
To detect hallucinations, self-consistency based methods have been proposed. For example, SelfCheckGPT~\citep{manakul2023selfcheckgpt} identifies hallucinated facts by measuring agreement across multiple sampled responses without external knowledge.  
Similarly, ~\citet{farquhar2024detecting} quantify uncertainty by computing entropy over semantically clustered outputs, revealing unstable and confabulated generations.  
Beyond these, verification-based approaches such as retrieval-augmented or fact-decomposition methods leverage external evidence to detect hallucinations, achieving strong performance but requiring additional supervision or tools~\citep{heo2025halucheck}.  

However, with the emergence of reasoning models, hallucinations occurring during multi-step reasoning have attracted increasing attention, as traditional hallucination detection methods that focus on final outputs often fail to capture errors embedded in long and complex reasoning chains~\citep{cheng2025chain}.  
~\citet{yao2025reasoning} show that reasoning models can be even more prone to hallucination in certain settings, especially due to miscalibrated uncertainty and post-training procedure. 
Other mechanistic perspectives analyze hallucination patterns unique to multi-step reasoning and propose tailored detection metrics and reward-shaping frameworks for reasoning paths~\citep{sun2025detection}.

\subsection{Reinforcement Learning for Hallucination Mitigation}
Reinforcement learning (RL) has recently been explored as an effective paradigm for hallucination mitigation by explicitly optimizing models toward factual and reliable behaviors.  
FSPO~\citep{lireasoning} introduces step-wise factuality rewards into policy optimization, encouraging correctness throughout the generation process rather than only at the final output.  
While RFT~\citep{song2025hallucination} leverages reinforcement learning to train models to appropriately refuse answering when uncertain, thereby reducing hallucinations by encouraging calibrated abstention.  
RLFH~\citep{wen2025policy} uses on-policy self-alignment with fine-grained feedback, decomposing responses into atomic statements and converting truth assessments into dense rewards.  
Additionally, TruthRL~\citep{wei2025truthrl} proposes a ternary reward design that explicitly distinguishes correct answers, hallucinations, and uncertainty, leading to significant reductions in hallucination rates across tasks.

However, existing approaches largely depend on external knowledge for detection and provide limited insight into how hallucinations arise within multi-step reasoning processes. Therefore, our work provides a fine-grained analysis of hallucinations during reasoning.
\section{Hallucination Analysis in the Reasoning Process of Large Language Models}
\label{sec:3}
\subsection{Preliminary}

For a large language model $\pi$ and a question $q$, we feed $q$ into the model and obtain a reasoning path with $m$ steps $ \mathcal{O} = {o_1, o_2, \ldots, o_m}$ and a final answer $A$. We assume a hallucination detector $H$, which determines whether a hallucination occurs in each reasoning step $o$:
\begin{align}
    H(o)\in\{0,1\}
\end{align}

Here, $H(o)=0$ indicates that a hallucination occurs, and $H(o)=1$ indicates otherwise. 

\subsection{Hallucination During Reasoning: Context-Sensitive Factual Hallucinations}
\label{sec:3.2}
Recently, as large language models (LLMs) continue to improve, the models have exhibited increasingly strong foundation capabilities; moreover, under the test-time scaling paradigm, they can tackle complex problems by allocating more inference-time compute and following longer reasoning trajectories~\citep{wu2024inference}.
However, hallucinations that arise within the reasoning steps can be subtle and may propagate across steps, making them harder to pinpoint than errors in short-form responses~\citep{yao2025reasoning, lireasoning}.  Moreover, while Chain-of-Thought prompting can reduce hallucination frequency in some cases, it can also obscure the cues relied upon by many existing hallucination detectors, thereby degrading detection effectiveness in long-reasoning settings~\citep{cheng2025chain}. 

Because the reasoning process of a large language model typically consists of multiple steps, we conduct a fine-grained analysis of whether hallucination occurs at each step. Consider a reasoning path $o_1, o_2, \ldots, o_m$; the model may hallucinate at step $o_k$, where this step contains a factual error.
However, if we rewrite the knowledge point corresponding to this hallucinated step into a standalone question and feed it to the model in isolation, as shown in Figure~\ref{fig:context-hallu}, the model can in fact produce a correct, non-hallucinated output:
\begin{align}
    H\left(\pi\left(o_k \mid o_{1:k-1}\right)\right)=0 \qquad
H\left(\pi\left(\hat o_k\right)\right)=1.
\end{align}

Here, $\hat o_k$ denotes the reformulation of the original $o_k$ into an independent question that is fed into the model in isolation. When $o_k$ is influenced by the context $o_{1:k-1}$, the model produces a hallucination; however, when $\hat o_k$ is provided independently without contextual influence, the model outputs the correct answer. This phenomenon indicates that the model actually possesses the relevant knowledge, but due to the influence of the reasoning context, it hallucinates during the reasoning process. This phenomenon is very common in hallucinations of reasoning models (see Section~\ref{sec:3.3} for a detailed empirical analysis).

Thus, we define this phenomenon as follows:

\paragraph{Definition }\textit{Context-sensitive factual hallucinations}:  the factual hallucination that arises during the model’s reasoning process, wherein the model actually possesses the relevant knowledge but, due to contextual influences during reasoning, produces a factually incorrect output.

\begin{figure}[t]
    \vspace{-9pt}
  \centering

  \begin{subfigure}[b]{0.49\textwidth}
    \begin{minipage}[t]{0.49\textwidth}
      \centering
      \includegraphics[width=\linewidth]{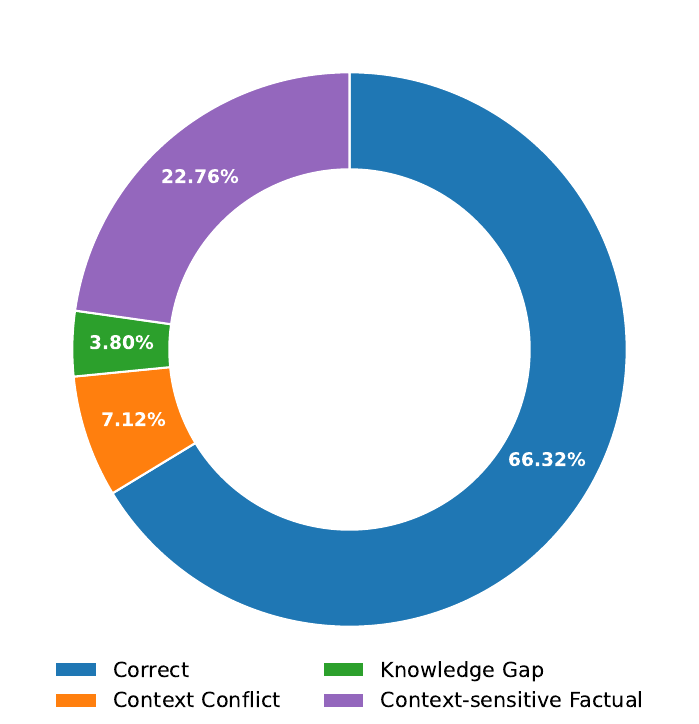}
      \caption*{\textbf{Total Distribution}}
    \end{minipage}%
    \begin{minipage}[t]{0.49\textwidth}
      \centering
      \includegraphics[width=\linewidth]{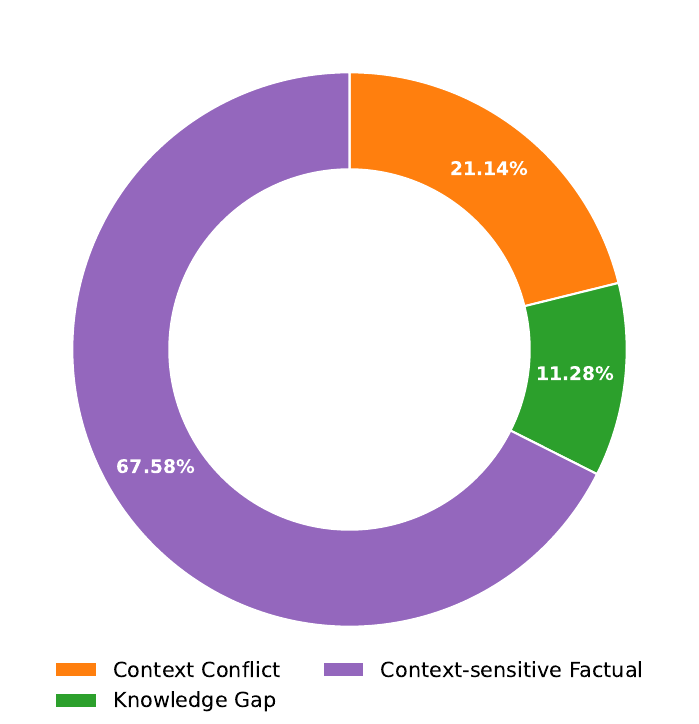}
      \caption*{\textbf{Hallucination Distribution}}
    \end{minipage}
    \caption{Qwen3-4B-Instruct}
    \label{fig:leftgroup2}
  \end{subfigure}
  \\
  \begin{subfigure}[b]{0.49\textwidth}
    \begin{minipage}[t]{0.49\textwidth}
      \centering
      \includegraphics[width=\linewidth]{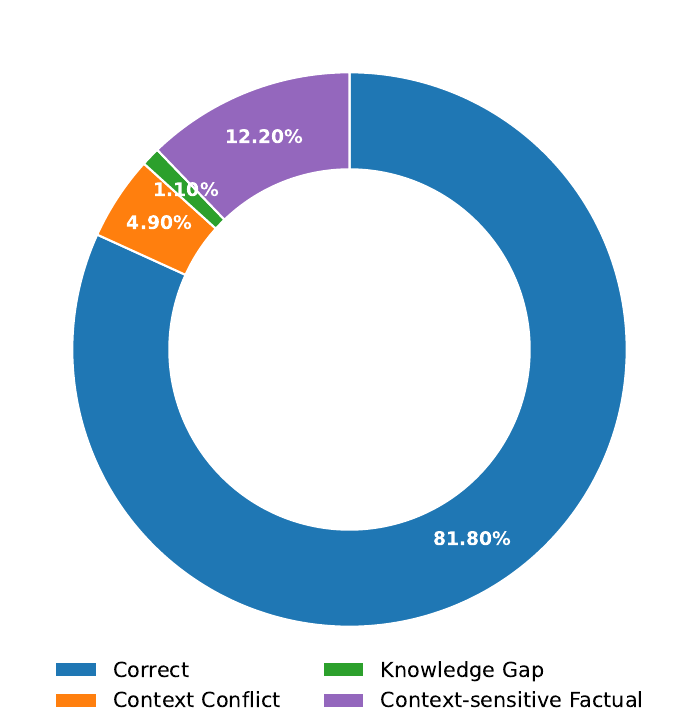}
      \caption*{\textbf{Total Distribution}}
    \end{minipage}%
    \begin{minipage}[t]{0.49\textwidth}
      \centering
      \includegraphics[width=\linewidth]{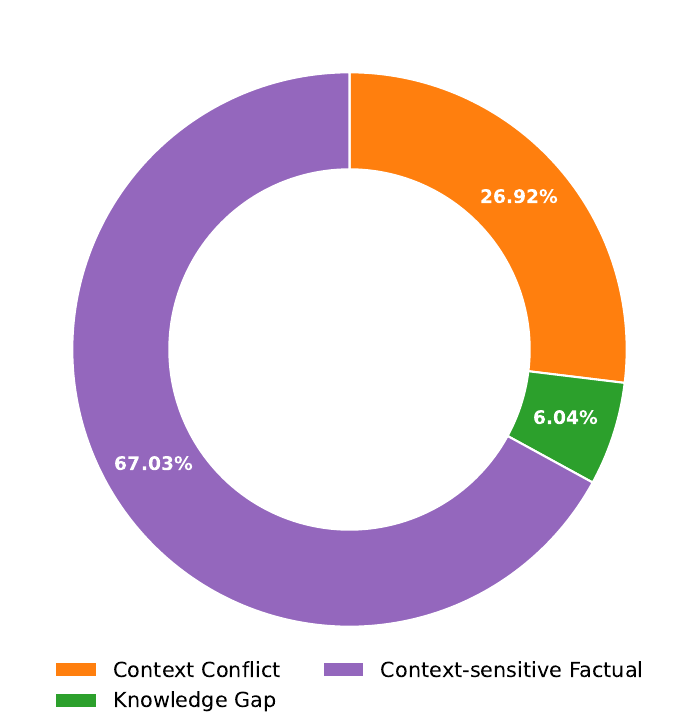}
      \caption*{\textbf{Hallucination Distribution}}
    \end{minipage}
    \caption{Qwen3-4B-Thinking}
    \label{fig:rightgroup2}
  \end{subfigure}

  \caption{\textbf{Analysis of hallucination types in reasoning models.}  Among all cases where hallucinations occur, context-sensitive factual hallucinations account for the majority—nearly 70\%. The other categories of hallucinations make up only a small fraction. As model capability improves, knowledge-missing hallucinations become increasingly rare, yet a substantial number of context-sensitive factual hallucinations still persist.}
   \vspace{-15pt}
  \label{fig:hallu_analysis}
\end{figure}

\begin{figure*}[th]
    \centering
  \includegraphics[width=0.88\linewidth]{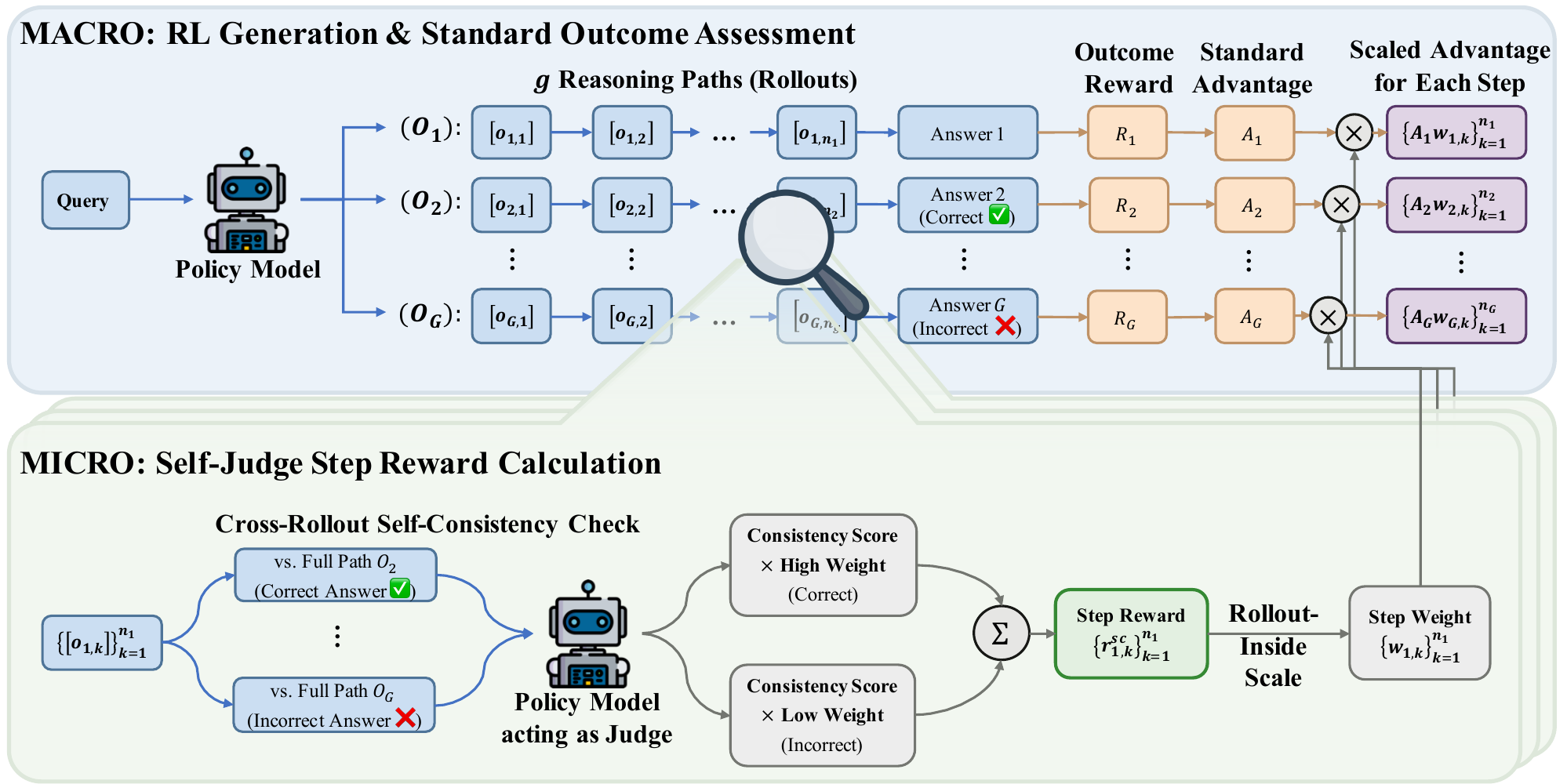}
  \caption {\textbf{Overview of Our Method.} We first follow the standard GRPO procedure to sample a group of rollouts. Then, for each reasoning step in each rollout, SSC-GRPO computes a consistency score with respect to other rollouts, which serves as a step-level reward. Based on these step-level rewards, we rescale the rollout-level advantages by allocating higher advantages to reasoning steps with higher consistency scores. Finally, SSC-GRPO optimizes the policy model using the rescaled advantages.}
  \label{fig:framework}
  \vspace{-10pt}
\end{figure*}

\subsection{Empirical Analysis of Reasoning-Process Hallucinations}
\label{sec:3.3}
To further validate the hypothesis above, we design an experiment to capture this phenomenon. The experimental procedure is as follows:

\begin{enumerate}
    \item Given a question $q$, we query the model to obtain a reasoning trace $\mathcal{O}=\{o_1,o_2,\cdots, o_m\}$ and corresponding answer $\mathcal{A}$.
    
    \item Then we use a stronger reference model to judge whether each answer $\mathcal{A}$ is correct. If $\mathcal{A}$ is correct, we label it as \textbf{correct}.
    
    \item If $\mathcal{A}$ is incorrect, we ask the reference model to locate the hallucinated step $o_k$ in the reasoning trace $\mathcal{O}$ and classify the error type as either \textbf{context conflict} or \textbf{factual hallucination}.
    
    \item If the error is a \textbf{factual hallucination}, we ask the reference model to rewrite the hallucinated fact at step $o_k$ into a standalone question, and then query the target model with this question in isolation. 
    If the model still fails to answer correctly, we attribute the error to a \textbf{knowledge gap}. 
    Otherwise, if the model answers correctly in isolation, we conclude that the original error is a \textbf{context-sensitive factual hallucination}.
\end{enumerate}

Through the steps above, we categorize hallucinations in the model's reasoning process into three types: \textbf{Context conflict}, \textbf{Knowledge gap}, and \textbf{Context-sensitive factual}. In practice, we evaluate Qwen3-4B-Instruct~\citep{yang2025qwen3} and Qwen3-4B-Thinking by sampling 2500 examples from the BOBA~\citep{AReaL-boba-Data} dataset and using Qwen3-235B-A22B as the reference model.

The results are shown in Figure~\ref{fig:hallu_analysis}. We observe that among the hallucinated samples, context-sensitive factual hallucinations account for nearly 70\%. As model capabilities improve, hallucinations caused by knowledge gaps become less frequent. This trend is especially pronounced for reasoning models such as Qwen3-4B-Thinking, whose accuracy is substantially higher than that of standard instruct models, resulting in very few knowledge-gap hallucinations. Nevertheless, a large number of context-sensitive factual hallucinations still remain.

Moreover, to further substantiate the conclusions above, we design a set of contextual data evaluation experiments in Appendix~\ref{sec:3.4}. The results still validate the correctness of our hypothesis.

\section{Method: Step-Level Self-Consistency Group Relative Policy Optimization}
To address the hallucination issue described above, we propose our method, \textbf{S}tep-level \textbf{S}elf-\textbf{C}onsistency \textbf{G}roup \textbf{R}elative \textbf{P}olicy \textbf{O}ptimization (SSC-GRPO). An overview of our approach is illustrated in Figure~\ref{fig:framework}.

\subsection{Background: GRPO}
Group Relative Policy Optimization (GRPO) is an RL algorithm proposed by DeepseekMath~\citep{shao2024deepseekmath}, aiming to estimate advantages without training a separate critic model. For each prompt $q$, GRPO samples a group of outputs from the old policy $\pi_{\text{old}}$ and assigns each output a reward score $\{R_i\}$. GRPO then constructs the advantage using within-group relative rewards, where the advantage is defined as:
\begin{equation}
A_i =\frac{R_i-\mathrm{mean}(\{R_i\}_{i=1}^G)}{\mathrm{std}(\{R_i\}_{i=1}^G)}.
\label{eq:grpo-adv}
\end{equation}
GRPO then performs policy updates using a PPO-style clipped surrogate objective, together with a KL regularization term to a reference policy $\pi_{\text{ref}}$ for stability:
\begin{equation}
\begin{aligned}
J_{\mathrm{GRPO}}(\theta)=
\mathbb{E}_{q,\{o_i\}\sim\pi_{\theta_{\text{old}}}}
\Bigg[
\frac{1}{G}\sum_{i=1}^{G}
\min\Big(
\rho_i(\theta)\,A_i,\;
\\
\mathrm{clip}(\rho_i(\theta),1-\epsilon,1+\epsilon)\,A_i
\Big)
\;-\;\beta\,D_{\mathrm{KL}}(\pi_\theta\|\pi_{\text{ref}})
\Bigg].
\end{aligned}
\label{eq:grpo-obj}
\end{equation}
where $\rho_i(\theta)=\frac{\pi_\theta(o_i\mid q)}{\pi_{\theta_{\text{old}}}(o_i\mid q)}$, $\epsilon$ is the clipping parameter, and $\beta$ controls the KL penalty. 

\subsection{Self-Consistency-Based Step-Level Reward}
\label{sec:4.2}
In standard GRPO, for a given input, the policy is rolled out multiple times to obtain a group of sampled outputs. Meanwhile, self-consistency-based hallucination detection methods likewise rely on multiple stochastic generations from the model and quantify agreement across these samples to estimate consistency. Therefore, the GRPO training setting is highly compatible with self-consistency. Moreover, as discussed in Section~\ref{sec:3}, hallucinations that occur during multi-step reasoning are often \emph{context-sensitive factual hallucinations}: the model actually possesses the relevant knowledge, yet produces factual errors due to contextual interference. This property enables us to detect such hallucinated segments \emph{without relying on external knowledge sources}. So, by leveraging self-consistency signals across rollouts, we can effectively identify the steps that exhibit context-sensitive factual hallucinations and consequently assign step-level rewards to guide policy optimization.

Specifically, for each rollout $o_i$, we decompose it into $n_i$ reasoning steps $\{o_{i,1},\ldots,o_{i,n_i}\}$. 
For each step $o_{i,k}$, we compute its self-consistency score against the reasoning parts of other rollouts in the same group (excluding the final answer).
Following the self-consistency checking scheme in SelfCheckGPT~\citep{manakul2023selfcheckgpt}, we use an NLI-style prompt to ask the training model $\pi_{old}$ to determine whether the step is \emph{entailment}, \emph{contradiction}, or \emph{neutral} with respect to another rollout:
\begin{equation}
\\SC_{i,k\leftarrow j}\;=\;\mathrm{NLI}_{\pi_{old}}\left(o_{j},\ o_{i,k}\right)
\in\{+1,0,-1\},
\label{eq:nli_label}
\end{equation}
where $+1$, $0$, and $-1$ correspond to \texttt{entailment}, \texttt{neutral}, and \texttt{contradiction}, respectively. The full prompt can be found in the Appendix~\ref{apdx:4}.

After obtaining the consistency signals against all other rollouts, we compute the final step-level consistency reward by a weighted aggregation:
\begin{equation}
r^{\mathrm{sc}}_{i,k}
\;=\;
\sum_{j\neq i} \alpha_j \cdot SC_{i,k\leftarrow j}.
\label{eq:step_sc_reward}
\end{equation}
Here, the weight $\alpha_j$ is determined by the outcome reward of the referenced rollout $o_j$.
If $R_j>0$, the referenced rollout is considered correct and thus receives a higher weight; 
if $R_j\le 0$, it is considered incorrect and receives a lower weight:
\begin{equation}
\alpha_j=
\begin{cases}
\alpha_{+}, & R_j>0\\
\alpha_{-}, & R_j\le 0
\end{cases}
 \quad \text{where} \quad \alpha_{+}>\alpha_{-}.
\label{eq:weight_by_outcome}
\end{equation}

\subsection{Step Advantage Rescaling}

After obtaining the step-level self-consistency rewards, we incorporate them into GRPO by re-scaling the advantage of each step according to $r^{\mathrm{sc}}_{i,k}$. Inspired by the method in~\citet{zhang2025criticsearch}, first, for each rollout, we compute a trajectory-level reward using the reward model, and then derive the trajectory-level advantage for each $o_i$ following Equation~\ref{eq:grpo-adv}. Next, within each rollout, we perform a step-level advantage redistribution by computing a per-step advantage coefficient:

\begin{equation}
\begin{aligned}
    w_{i,k} = \mathrm{clip}\Big(1 + \mathrm{sgn}(A_i)\lambda \frac{r^{\mathrm{sc}}_{i,k} - \mathrm{mean}(\{r^{\mathrm{sc}}_{i,k}\}_{k=1}^{n_i})}{\mathrm{std}(\{r^{\mathrm{sc}}_{i,k}\}_{k=1}^{n_i})}\\,
    1-\epsilon_c,1+\epsilon_c \Big)
    \end{aligned}
    \label{eq:scaled}
\end{equation}

Here, $\mathrm{sgn}(\cdot)$ denotes the sign function. When $A_i$ is positive, sentences with higher consistency receive larger weights; when $A_i$ is negative, sentences with higher consistency receive smaller weights, so that reasoning steps with larger $r^{\mathrm{sc}}_{i,k}$ are assigned larger advantages. 
The parameter $\lambda$ is the scaling factor and $\epsilon_c$ is the clipping coefficient that prevents excessively large reallocation magnitudes from harming training stability. 
With Equation~\ref{eq:scaled}, we use centered step-level consistency scores to construct clipped multiplicative weights for the rollout-level advantages.

\subsection{Reinforcement Learning Formulation}
We now formalize the overall training objective of \textbf{SSC-GRPO}.
Given a reference model $\pi_{old}$ and a reference model $\pi_{ref}$, based on $G$ rollouts $\{o_i\}_{i=1}^G$ for input $q$, the objective of \textbf{SSC-GRPO} is to optimize the policy $\pi_\theta$ by maximizing the following objective:

\begin{equation}
\begin{aligned}
J_{\mathrm{SSC-GRPO}}(\theta)=
\mathbb{E}_{q,\{o_i\}\sim\pi_{\theta_{\text{old}}}}
\Bigg[
\frac{1}{G}\sum_{i=1}^{G}\frac{1}{T_i}\sum_{t=1}^{T_i} \\
\min\Big( 
\rho_{i,t}(\theta)\,A_i\,w_{i,k(t)},\; 
\mathrm{clip}\!\big(\rho_{i,t}(\theta),1-\epsilon,1+\epsilon\big) \\ \,A_i\,w_{i,k(t)}
\Big) 
\;-\;\beta\,D_{\mathrm{KL}}(\pi_\theta\|\pi_{\text{ref}})
\Bigg],
\end{aligned}
\label{eq:sc-grpo-obj}
\end{equation}
where $k(t)$ denotes the index of the reasoning step to which token $t$ belongs, and $\rho_{i,t}(\theta)=\frac{\pi_\theta(o_{i,t}\mid q,o_{i,<t})}{\pi_{\theta_{\text{old}}}(o_{i,t}\mid q,o_{i,<t})}$.

\section{Experiment}
In this section, we describe our experimental setup, report the main results, and provide further analysis.
\begin{table*}[h]
    \centering
        
         \resizebox{\linewidth}{!}{
        \begin{tabular}{cccccccccc}
\toprule[1.2pt]
\multirow{2}{*}{\textbf{Model}} & \multirow{2}{*}{\textbf{Method}}  & \multicolumn{4}{c}{\textbf{Math}} & \multicolumn{3}{c}{\textbf{Hallucination}} & \multicolumn{1}{c}{\multirow{2}{*}{\textbf{Avg.}}} \\
\cmidrule(lr){3-6} \cmidrule(lr){7-9} 
& & GSM8K & MATH-500 & AIME24 & AIME25 & HaluEval & TruthFulQA & HotpotQA &\\

\midrule

\multirow{5}{*}{Qwen3-4B-Base} & Vanilla  & 87.60 & 54.40 & 16.67 & 16.67 & 48.00 & 18.54 & 9.06 & 35.85 \\

& GRPO                             & \textbf{92.64} & \underline{73.40} & 33.33 & 26.67 & 51.08 & \underline{19.58} & 9.72 & 43.77 \\
& FSPO                             & 90.98 & 72.00 & 30.00 & 30.00 & \textbf{54.32} & 18.72 & \textbf{10.94} & 43.85 \\
& SEED-GRPO                             & 90.44 & 73.00 & \underline{36.67} & \underline{36.67} & \underline{53.30} & 18.24 & 9.94 & \underline{45.46} \\
& \textbf{SSC-GRPO}                            & \underline{92.03} & \textbf{74.60} & \textbf{36.67} & \textbf{43.33} & 52.42 & \textbf{21.17} & \underline{10.58} & \textbf{47.26} \\
\midrule
\multirow{5}{*}{Qwen3-4B-Ins} & Vanilla  & 85.67 & 78.00 & 60.00 & 53.33 & 65.00 & 45.90 & 16.40 & 57.75 \\

& GRPO                             & 89.76 & \underline{84.80} & 73.33 & 60.00 & 69.80 & \underline{54.46} & 17.80 & 64.27 \\
& FSPO                             & \underline{90.14} & 83.60 & 70.00 & 56.67 & \textbf{73.42} & 49.08 & \underline{18.73} & 63.09 \\
& SEED-GRPO                             & 89.84 & 85.00 & \underline{73.33} & \underline{66.67}  & 69.26 & 52.26 & 18.54 & \underline{64.98} \\
& \textbf{SSC-GRPO}                             & \textbf{90.60} & \textbf{85.40} & \textbf{73.33} & \textbf{66.67} & \underline{71.12} & \textbf{55.32} & \textbf{19.46} & \textbf{65.99} \\
\midrule
\multirow{5}{*}{Llama3-8B-Ins} & Vanilla  & 81.20 & 44.60 & 16.67 & 0.00 & 65.34 & 35.50 & 15.72 & 37.00 \\

& GRPO                             & \underline{84.91} & 54.00 &23.33 & 6.67 & \underline{69.80} & 40.27 & 23.78 & 43.25 \\
& FSPO                             & 83.99 & 52.80 & 23.33 & \underline{10.00} & 68.90 & \underline{41.13} & \textbf{25.22} & 43.62 \\
& SEED-GRPO                             & \textbf{85.28} & \textbf{55.60} & \underline{26.67} & 6.67 & 69.40 & 40.64 & 23.88 & \underline{44.01} \\
& \textbf{SSC-GRPO}                             & 84.82 & \underline{55.20} & \textbf{30.00} & \textbf{10.00} & \textbf{71.36} & \textbf{42.84} & \underline{24.34} & \textbf{45.51} \\
\bottomrule[1.2pt]

\end{tabular}} %
    
    \caption{\textbf{The Main results of our methods in math and hallucination benchmarks.} For AIME24 and AIME25, we report Pass@32 accuracy, while Pass@1 accuracy is used for the other benchmarks. The best results are highlighted in \textbf{bold} and the second best results are highlighted in \underline{underline}. The SSC-GRPO method achieves state-of-the-art performance on Qwen3-4B Base, Qwen3-4B-Instruct and Llama3-8B-Instruct, outperforming the baseline by an average of 1.80\%, 1.01\% and 1.50\%, respectively.}
    \label{tab:cpt_result}
    \vspace{-15pt}
\end{table*}
\subsection{Experimental Setup}

\paragraph{Datasets and Benchmarks.}
Since our goal is to mitigate hallucinations in long reasoning while preserving general capabilities, we evaluate on both mathematical reasoning and hallucination benchmarks. Overall, our experimental protocol largely follows the setup in FSPO~\citep{lireasoning}. For hallucination-oriented training, we use a 2K subset of HotpotQA~\citep{yang2018hotpotqa} as the training set. For mathematical training, we mix 200 high-quality samples from the BOBA~\citep{AReaL-boba-Data} dataset with 8,000 samples from the Simple-RL~\citep{zeng2025simplerlzooinvestigatingtamingzero} dataset.

For evaluation, we test math on GSM8K~\citep{cobbe2021gsm8k}, MATH-500~\citep{hendrycks2021measuring}, AIME-24~\citep{aime24}, and AIME-25~\citep{aime25}. We report Pass@1 accuracy for GSM8K and MATH-500. For AIME-24 and AIME-25, since Pass@1 accuracy can exhibit high variance at test time, we report Pass@32 accuracy for improved stability. For hallucination-related evaluation, we validate on HaluEval(QA task)~\citep{li2023halueval}, TruthfulQA~\citep{lin2022truthfulqa}, and the test set of HotpotQA~\citep{yang2018hotpotqa}. All evaluations were conducted twice, and we report the average results.
.

\paragraph{Models.}
We choose three base models: Qwen3-4B-base~\citep{yang2025qwen3}, Qwen3-4B-Instruct~\citep{yang2025qwen3} and Llama3-8B-Instruct~\citep{grattafiori2024llama}. On top of each base model, we apply our method as well as competing baselines.

\paragraph{Baselines.}
We compare SSC-GRPO against the following methods: \textbf{Vanilla}: the original base model without RL fine-tuning; \textbf{GRPO}~\citep{shao2024deepseekmath}: the standard group relative policy optimization algorithm; \textbf{FSPO}~\citep{lireasoning}: the method factuality-aware step-wise policy optimization, which assigns step-level rewards using external knowledge to guide policy optimization; \textbf{SEED-GRPO}~\citep{chen2025seed}: which reweights advantages across trajectories using semantic entropy.
All the aforementioned methods are reproduced under identical settings on the same dataset, ensuring a consistent experimental environment.
\paragraph{Implementation Details.}
We implement and train all methods using the verl~\citep{sheng2024hybridflow} framework. The full hyperparameter settings are provided in Appendix~\ref{apdx:1}.

\begin{table}[t]
    \centering
        
         \resizebox{\linewidth}{!}{
        
\begin{tabular}{ccccc}
\toprule[1.2pt]
\textbf{Method} & \textbf{Correct} & \textbf{Knowledge Gap} & \textbf{Context Conflict} & \textbf{Context Sensitive}\\
\midrule

Qwen3-4B-Ins & 66.32\% & 3.80\% & 7.12\% & 22.76\% \\
SSC-GRPO& 72.24\% & 3.18\% & 7.68\% & 16.90\%  \\
\bottomrule[1.2pt]
\end{tabular}

} %
    \caption{\textbf{Hallucination type after training.} After applying SSC-GRPO, the proportion of context-sensitive factual hallucinations decreases significantly.}
    \vspace{-15pt}
    \label{tab:ssc-impact}
\end{table}

\subsection{Main Results}
Our main experimental results are shown in Table~\ref{tab:cpt_result}. On Qwen3-4B-Base, Qwen3-4B-Instruct and Llama3-8B-Instruct, our method, SSC-GRPO, consistently outperforms the baseline and other competing approaches, achieving state-of-the-art performance. On Qwen3-4B-Base, SSC-GRPO improves over the baseline by 1.80\% on average. On Qwen3-4B-Instruct SSC-GRPO exceeds the baseline by 1.01\%, while on Llama3-8B-Instruct, it exceeds the baseline by 1.50\%.

Although SEED-GRPO ties our method on AIME24 and AIME25 with Qwen3-4B-Instruct, we still achieve better average performance across mathematical benchmarks. In particular, on Qwen3-4B-Base, our method surpasses other approaches by 6.67\%. 
For hallucination benchmarks, FSPO achieves higher performance than SSC-GRPO on HaluEval; however, it relies on external knowledge bases to provide reward signals. In contrast, our method requires only the model’s internal knowledge and still outperforms other approaches on the overall hallucination benchmarks. These results demonstrate the effectiveness of SSC-GRPO.
\subsection{Can SSC-GRPO mitigate context-sensitive factual hallucination?}
In Section~\ref{sec:3}, we pointed out that most existing hallucinations are context-sensitive factual hallucinations. To verify whether our method can effectively mitigate this type of hallucination, we applied the hallucination classification method from Section~\ref{sec:3.3} to the model after SSC-GRPO training. The results are shown in Table~\ref{tab:ssc-impact}.

We observe that after applying our method, the proportion of context-sensitive factual hallucinations decreases significantly, indicating that SSC-GRPO can effectively mitigate this issue.

\begin{figure}[t]
    \centering
  \includegraphics[width=0.9\linewidth]{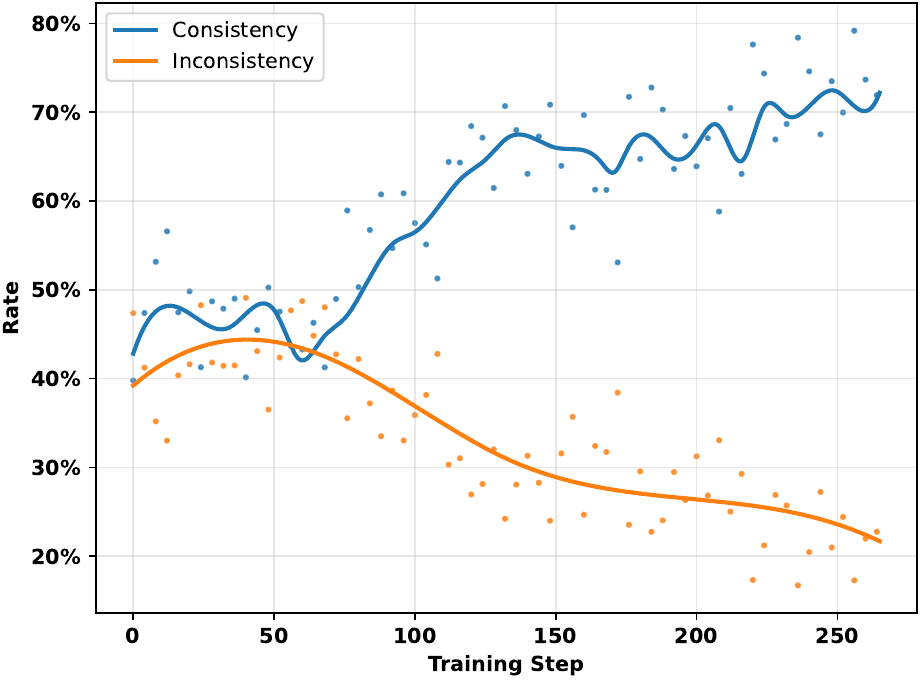}
  \caption {\textbf{Consistency Ratio over Training.}
As training progresses, the proportion of consistent steps steadily increases while the proportion of inconsistent steps decreases, demonstrating that hallucinations are gradually reduced during training.}
  \label{fig:sc_rate}
\end{figure}

\begin{table}[t]
    \centering
        \small
         \resizebox{\linewidth}{!}{
        
\begin{tabular}{ccccc}
\toprule[1.2pt]
\textbf{Model} & \textbf{Case Number} & \textbf{Accuracy} & \makecell{\textbf{Entailment}  \\\textbf{Precision}} & \makecell{\textbf{Contradiction}  \\\textbf{Precision}}  \\
\midrule
Qwen3-4B-Base & 210 & 74.76 & 95.40 & 70.96 \\
\addlinespace[1pt]
Qwen3-4B-Ins & 553 & 84.62 & 97.40 & 76.27 \\
\bottomrule[1.2pt]
\end{tabular}

} %
            \caption{\textbf{Evaluation of the model’s NLI capability.} We use GPT-4.1’s results as ground truth. Both Qwen3-4B Base and Qwen3-4B-Instruct achieve high accuracy and precision on NLI task, demonstrating that the models possess strong NLI reasoning capabilities.}
    \vspace{-10pt}
    \label{tab:rollnli}
\end{table}

\subsection{Step Consistency Over the Process of Training}

During training, we compute the consistency of each reasoning step with respect to other rollouts. We expect that, as training progresses, reasoning consistency will increase, thereby reducing hallucinations. To verify this hypothesis, we track the proportion of consistent and inconsistent steps over the course of training. 

As shown in Figure~\ref{fig:sc_rate}, the proportion of consistent steps gradually increases while the proportion of inconsistent steps decreases. This indicates that reinforcement learning progressively enhances reasoning consistency and reduces hallucinations, which in turn improves the model’s reasoning capability and leads to performance gains across benchmarks.

\subsection{Accuracy of Self-consistency Judge}
\label{sec:6.1}



SSC-GRPO relies on the model’s NLI judgments to construct step-level consistency signals. To assess this capability on rollout-derived reasoning, we initially sample 20 training questions for each model and construct premise–hypothesis pairs using the cross-rollout procedure and NLI prompt described in our method. After filtering invalid rollout groups, we retain 210 pairs for Qwen3-4B-Base and 553 for Qwen3-4B-Instruct. Since these pairs lack ground-truth NLI annotations, we use GPT-4.1 as an independent reference judge and compare its labels with the corresponding model’s self-judgments.

As shown in Table~\ref{tab:rollnli}, Base and Instruct achieve overall accuracy rates of 74.76\% and 84.62\%, respectively. Both models achieve high precision for entailment and contradiction. Since our method requires high precision, missed detections merely leave training equivalent to standard GRPO. Therefore, high precision indicates that our NLI signals are effective. This indicates that the models possess strong NLI discrimination capability and also supports the reasonableness of using the model itself for consistency judgment. Full confusion matrices and evaluation details are provided in Appendix~\ref{apdx:5}.

\section{Conclusion}

In this paper, we conduct a fine-grained analysis of hallucinations in the reasoning process and identify \emph{context-sensitive factual hallucination}: the model actually possesses the relevant knowledge, yet produces factual errors due to interference from the reasoning context. Through empirical analysis, we show that this type of hallucination accounts for the majority of hallucinations during reasoning. To address this issue, we propose \textbf{S}tep-Level \textbf{S}elf-\textbf{C}onsistency \textbf{G}roup \textbf{R}elative \textbf{P}olicy \textbf{O}ptimization(SSC-GRPO), which leverages self-consistency to provide step-level rewards for reasoning trajectories. Experiments on both mathematical and hallucination benchmarks demonstrate the effectiveness of \textsc{SSC-GRPO}. Our work offers a new perspective on mitigating hallucinations in the reasoning process.

\section*{Limitations}

Due to resource constraints, our evaluation focuses on 4B- and 8B-scale models from the Qwen and Llama families on mathematical reasoning and hallucination-related benchmarks. Further evaluation is needed to assess the applicability of SSC-GRPO to larger models and a broader range of tasks. In addition, cross-rollout consistency checking introduces extra training cost, as quantified in Appendix~\ref{apdx:c}. Future work could explore selective consistency checking and more efficient inference to reduce this overhead, while examining their effects on the quality of the consistency signals.

\section*{Ethical Considerations}
We utilized generative AI to facilitate code debugging and to refine the writing style with grammatical errors.

\bibliography{custom}

\appendix

\newtcolorbox{promptbox}[2][]{
  width=\linewidth,
  boxrule=1.5pt,
  colframe=black,                 
  colback=gray!15,               
  arc=5pt,
  rounded corners,
  left=6pt, right=6pt, top=6pt, bottom=6pt,  
  before skip=6pt, after skip=6pt,
  title={#2},
  fontupper=\scriptsize,        
  fonttitle=\ttfamily\bfseries\color{white}, 
  colbacktitle=black,            
  frame hidden=false,
  boxrule=1.5pt,
  titlerule=0pt,                 
  top=3pt,
  #1
}
\section{Implementation Details of our method}
\label{apdx:1}
The detailed hyperparameter settings are shown in Table~\ref{tab:inp}. Our training is conducted on 8 Nvidia H20 GPUs, with full parameter optimization and gradient checkpointing. 

\begin{table}[h]
    \centering
    \small
        
         \resizebox{0.8\linewidth}{!}{
        
\begin{tabular}{c|c}
\toprule[1.2pt]
\textbf{Parameter} & \textbf{Value} \\
\midrule 
Training batch size & 32 \\
Learning rate & 1e-6 \\
Training epochs & 2\\
Rollout number & 8 \\
Rollout temperature &1.0 \\
KL loss coefficient $\beta$ & 0.001 \\
Clip ratio & 0.2 \\
Positive reward factor $\alpha_+$ & 0.8 \\
Negative reward factor $\alpha_-$ & 0.2 \\
Rescaled factor $\lambda$ & 0.05 \\
Rescaled clip ratio $\epsilon_c$ & 0.2 \\
\bottomrule[1.2pt]
\end{tabular}

} %
    
    \caption{\textbf{The hyperparameter of our experiments}}
    \label{tab:inp}
\end{table}

\section{Effect of the Factor $\lambda$ and $ \epsilon_c $}

In Equation~\ref{eq:scaled}, we apply both scaling and clipping to the coefficient $w$. To examine the impact of this coefficient, we conduct ablation studies over different values of $\lambda$ and $\epsilon_c$. Specifically, we train the Qwen3-4B-Instruct model on a mathematical dataset using various parameter settings, and the results are shown in Table~\ref{tab:abl}. We find that removing scaling and clipping causes the training process to collapse. Larger values of $\lambda$ or smaller values of $\epsilon_c$ also negatively affect training stability. Therefore, we ultimately choose $\lambda = 0.05$ and $\epsilon_c = 0.2$ as the final hyperparameters.

\begin{table}[t]
    \centering
        
         \resizebox{\linewidth}{!}{
        
\begin{tabular}{ccccc}
\toprule[1.2pt]
\textbf{Parameter} & GSM8K & Math-500 & AIME24 &AIME25 \\
\midrule
w/o $\lambda$ and $\epsilon_c$ & 85.06 & 76.00 & 60.00 & 50.00 \\

$\lambda=0.2$,$\epsilon_c = 0.2$  & 88.86 & 82.80 & 66.67 & 60.00 \\
$\lambda = 0.05$, $\epsilon_c = 0.1$ & 90.52 & 83.60 & 70.00 & 63.33 \\
$\lambda = 0.05$, $\epsilon_c = 0.2$ & \textbf{90.60} & \textbf{85.40} & \textbf{73.33} & \textbf{66.67} \\
\bottomrule[1.2pt]
\end{tabular}

} %
        \caption{\textbf{Ablation study of the factor $\lambda$ and $\epsilon_c$ in Qwen3-4B-Instruct.}  We choose $\lambda = 0.05$ and $\epsilon_c = 0.2$ as the final hyperparameters.}

    \label{tab:abl}
\end{table}

\section{Computational Cost Evaluation for SSC-GRPO}
\label{apdx:c}
SSC-GRPO requires self-consistency evaluation, which introduces additional computational overhead. We measure its time cost relative to standard GRPO across different models and quantify the proportion of time spent on NLI inference. We run ten training steps on eight H20 GPUs and report the average time per step in Table~\ref{tab:time}.

\begin{table}[t]
    \centering
        
         \resizebox{\linewidth}{!}{
        
\begin{tabular}{ccccc}
\toprule[1.2pt]
\textbf{Model} & \textbf{GRPO seconds} & \textbf{SSC-GRPO seconds} & \textbf{NLI judge seconds} & \textbf{NLI Time Share}  \\
\midrule
Qwen3-4B-Base & 44.49 & 65.12  & 18.13 & 27.84\% \\
\addlinespace[1pt]
Qwen3-4B-Ins & 64.62 & 92.26 & 24.81 & 26.89\% \\
\bottomrule[1.2pt]
\end{tabular}

} %
        \caption{\textbf{Computational cost per step for SSC-GRPO} We ran ten training steps and averaged the time spent in each stage across these steps.}

    \label{tab:time}
\end{table}

Although our method introduces additional computational overhead, the NLI stage involves inference only and requires a single-word output, which helps limit its cost. NLI inference accounts for approximately 27\% of the total training time, while SSC-GRPO incurs approximately 40\% additional runtime compared with standard GRPO. Thus, although SSC-GRPO adds computational complexity, its practical runtime overhead remains manageable.

\section{Cross-Model Validation of Hallucination Classification}

In the main analysis, we use Qwen3-235B-A22B as the reference model to identify and classify hallucinations in reasoning traces. To examine whether the resulting taxonomy distribution is sensitive to the choice of the judge model, we repeat the analysis on the same sampled Qwen3-4B-Instruct reasoning traces using GPT-OSS-120B as an alternative judge.

\begin{table}[t]
    \centering
        
         \resizebox{\linewidth}{!}{
        
\begin{tabular}{lcccc}
\toprule[1.2pt]
\textbf{Judge Model}
& \textbf{Correct}
& \textbf{Context Conflict}
& \textbf{Knowledge Gap}
& \textbf{Context-Sens. Factual} \\
\midrule[1.2pt]
Qwen3-235B-A22B & 66.32\% & 7.12\% & 3.80\% & 22.76\% \\
GPT-OSS-120B    & 64.40\% & 10.20\% & 4.40\% & 21.00\% \\
\bottomrule[1.2pt]
\end{tabular}
} %
        \caption{
Hallucination taxonomy distributions produced by two judge models on the same sampled Qwen3-4B-Instruct reasoning traces. Both judges support the same qualitative conclusion.
}
\label{tab:cross_judge}
    
\end{table}

As shown in Table~\ref{tab:cross_judge}, the two judge models yield similar aggregate distributions. In particular, context-sensitive factual hallucinations account for 22.76\% and 21.00\% of all examples under Qwen3-235B-A22B and GPT-OSS-120B, respectively. Despite differences in individual category proportions, context-sensitive factual hallucination remains the most frequent error category under both judges. Therefore, using different large-scale judge models leads to the same qualitative observation, suggesting that the hallucination taxonomy analysis is robust to the choice of the reference model. This consistency across independent judge models provides additional evidence that LLM-based judging offers a meaningful and reliable approach to hallucination-type analysis.

\section{Further Analysis: Context-Augmented Data Evaluation}
\label{sec:3.4}
To further substantiate the conclusions above, we design a set of contextual data evaluation experiments. Since most existing benchmarks evaluate models on single, independent questions, we aim to better approximate scenarios that arise during multi-step reasoning. Inspired by the idea in~\citet{shi2023large}, we apply context-form data augmentation to the original questions: we append additional constraints from the same domain that do not affect the original solution, i.e., extra conditions that are irrelevant to answering the question. With such perturbations, the model's process of solving the question becomes analogous to solving intermediate sub-questions during a reasoning procedure. The augmented examples are shown in Appendix~\ref{apdx:2}.

In practice, we perform contextual data augmentation on subsets of GSM8K~\citep{cobbe2021gsm8k}, MATH~\citep{hendrycks2021measuring}, and DeepScaler~\citep{deepscaler2025}. Using a prompt (see Appendix~\ref{apdx:4}), we instruct Qwen3-235B-A22B~\citep{yang2025qwen3} to perturb the original questions according to our requirements while ensuring that the solution process and final answer remain unchanged. We then evaluate Qwen3-4B-Instruct and Qwen3-4B-Thinking on both the original benchmarks and the augmented versions, and compare their accuracies.

\begin{table}[h]
    \centering
        
         \resizebox{\linewidth}{!}{
        \begin{tabular}{ccccccc}
\toprule[1.2pt]
\multirow{2}{*}{\textbf{Model}} & \multicolumn{2}{c}{\textbf{GSM8K}} & \multicolumn{2}{c}{\textbf{Math-500}} & \multicolumn{2}{c}{\textbf{DeepscaleR}} \\
\cmidrule(lr){2-3} \cmidrule(lr){4-5}  \cmidrule(lr){6-7}

 & Origin & Augment & Origin & Augment & Origin & Augment \\
\midrule

\multirow{1}{*}{Qwen3-4B-Instruct}   & 89.69 & 75.13 & 78.00 & 71.20 & 59.60 & 51.20 \\
\addlinespace[1pt]
\rowcolor{gray!12}
$\Delta$ & \multicolumn{2}{c}{-14.56\%} &
\multicolumn{2}{c}{-6.80\%} &
\multicolumn{2}{c}{-8.40\%} \\
\midrule
\multirow{1}{*}{Qwen3-4B-Thinking}   & 94.76 & 81.50 & 86.20 & 80.20 & 63.90 & 57.70 \\
\rowcolor{gray!12}
\addlinespace[1pt]
$\Delta$ & \multicolumn{2}{c}{-13.26\%} &
\multicolumn{2}{c}{-6.00\%} &
\multicolumn{2}{c}{-6.20\%} \\
\bottomrule[1.2pt]
\end{tabular}} %
        \caption{\textbf{Results of Context-Augmented Data Evaluation.} After augmentation, both Qwen3-4B-Instruct and Qwen3-4B-Thinking show a significant performance drop. For the relatively simple GSM8K benchmark, the decline is the most severe, reaching 14\%. This indicates that once subjected to contextual perturbations, the models’ reasoning on the original questions is indeed affected. This corroborates the existence of context-sensitive factual hallucinations.}

    \label{tab:aug}
\end{table}
We observe that across all benchmarks, the accuracy on the augmented data drops by nearly 10\% relative to the original questions. On the simpler GSM8K benchmark, the degradation is even more pronounced. These results indicate that, for current models, irrelevant contextual information can indeed interfere with the reasoning needed to solve the original problem, providing further evidence for the existence of context-sensitive factual hallucinations.

\section{Example of Augmented Data}
\label{apdx:2}
\begin{figure}[h]
    \centering
  \includegraphics[width=0.9\linewidth]{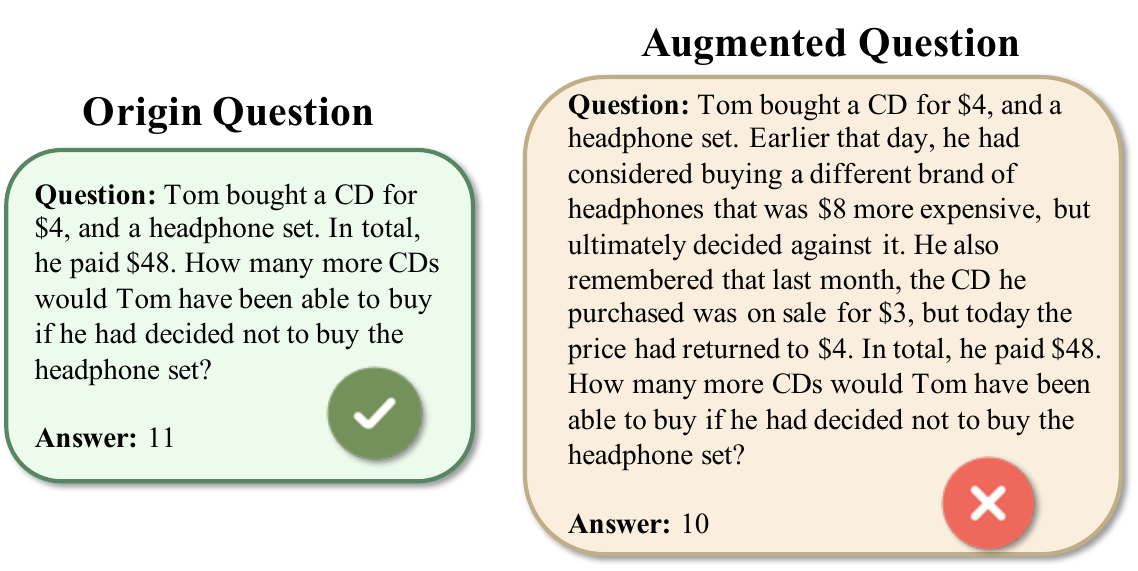}
  \caption {\textbf{Example of Augmented benchmarks.}For the original problem, the model produces the correct answer, 11. However, for the augmented version, the model is misled by the additional context and outputs 10, which is incorrect.
}
  \label{fig:example}
\end{figure}

In Section~\ref{sec:3.4}, we perform data augmentation on the original benchmark using prompts, introducing additional conditions that commonly appear in reasoning contexts but do not affect the final answer. Figure~\ref{fig:example} presents a comparison example before and after augmentation. Although extra conditions are added, these conditions are redundant and irrelevant to the actual solution. Nevertheless, the model produces an incorrect answer, indicating that it is indeed influenced by the reasoning context.

\section{SSC-GRPO Performance in Augmented Benchmarks}
In Section~\ref{sec:3.4}, we perform data augmentation on several benchmarks and observe that model performance drops significantly on the augmented datasets. We also evaluate SSC-GRPO on these augmented benchmarks, and the results are shown in the Table~\ref{tab:ssc}. Although performance still decreases after training with our method, SSC-GRPO consistently outperforms the original model and exhibits a smaller performance drop. Therefore, this experiment further provides indirect evidence of the effectiveness of our method.

\section{Confusion Matrix for Rollout NLI Judge}
\label{apdx:5}
\begin{figure}[h]
    \centering
  \includegraphics[width=1\linewidth]{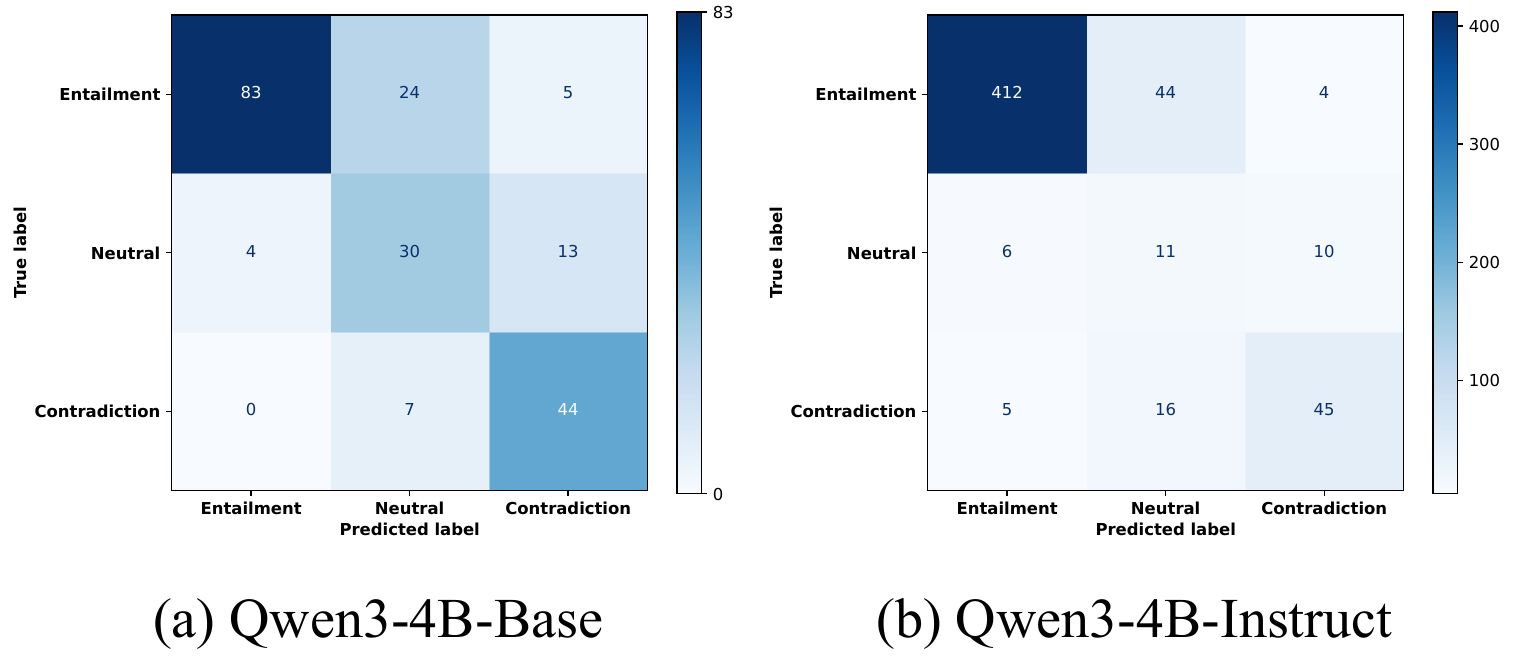}
  \caption {\textbf{The Confusion Matrix of rollout NLI task.} We use GPT-4.1’s results as ground truth.
}
  \label{fig:rol_conf}
\end{figure}

In Section~\ref{sec:6.1}, we report selected results of NLI judging on actual rollouts. The full confusion matrices are shown in Figure~\ref{fig:rol_conf}. Both models achieve high precision for entailment and contradiction. Most errors involve misclassifying examples as neutral; these cases do not affect training and simply retain the original GRPO behavior. These results provide further support for the conclusions in Section~\ref{sec:6.1}.

\begin{table}[h]
    \centering

         \resizebox{\linewidth}{!}{
        \begin{tabular}{ccccc}
\toprule[1.2pt]
\multirow{2}{*}{\textbf{Model}} & \multicolumn{2}{c}{\textbf{GSM8K}} & \multicolumn{2}{c}{\textbf{Math-500}}  \\
\cmidrule(lr){2-3} \cmidrule(lr){4-5}  

 & Origin & Augment & Origin & Augment \\
\midrule

\multirow{1}{*}{Qwen3-4B-Instruct}   & 89.69 & 75.13 & 78.00 & 71.20 \\
\addlinespace[1pt]
\rowcolor{gray!12}
$\Delta$ & \multicolumn{2}{c}{-14.56\%} &
\multicolumn{2}{c}{-6.80\%} \\
\midrule
\multirow{1}{*}{SSC-GRPO}   & 90.60 & 79.07 & 85.40 & 81.00  \\
\rowcolor{gray!12}
\addlinespace[1pt]
$\Delta$ & \multicolumn{2}{c}{-11.53\%} &
\multicolumn{2}{c}{-4.40\%} \\
\bottomrule[1.2pt]
\end{tabular}} %
    \caption{\textbf{Augmented benchmarks for our method}}
    
    \label{tab:ssc}
\end{table}

\begin{table}[t]
    \centering
        \small
         \resizebox{\linewidth}{!}{
        \begin{tabular}{cccc}
\toprule[1.2pt]
\textbf{Model} & \textbf{Accuracy} & \textbf{F1-score} & \textbf{Precision}\\
\midrule
Qwen3-4B-Base & 83.30 & 82.94 & 86.51\\

Qwen3-4B-Ins & 92.18 & 92.02 & 93.36\\
\bottomrule[1.2pt]
\end{tabular}} %
            \caption{\textbf{Evaluation of the model’s NLI capability on MathNLI.}}
    
    \label{tab:mathnli}
\end{table}

\begin{table}[t]
    \centering
        
         \resizebox{\linewidth}{!}{
        
\begin{tabular}{cccccc}
\toprule[1.2pt]
\textbf{Model} & \textbf{Acc} & \makecell{\textbf{Entailment}  \\\textbf{Precision}} & \makecell{\textbf{Contradiction}  \\\textbf{Precision}} & \makecell{\textbf{CER} \\ (\small$ +1\to-1, -1\rightarrow+1$)} & 
\makecell{\textbf{FAR} \\ ($0\rightarrow\pm 1$)} \\
\midrule
Qwen3-4B-Base & 54.50 & 94.79 & 94.81 & 0.25\% & 0.92\%\\
\addlinespace[1pt]
Qwen3-4B-Ins & 90.81 & 93.91 & 97.46 & 0.78\% & 2.02\%\\
\bottomrule[1.2pt]
\end{tabular}

} %
    \caption{\textbf{Evaluation of the model’s NLI capability on SNLI.}  Critical Error Rate (CER) measures entailment–contradiction label reversals, while False Activation Rate (FAR) measures cases whose true label is neutral but whose predicted label is entailment or contradiction.}
    
    \label{tab:nli}
\end{table}

\section{Extra results of NLI Tasks}
\label{apdx:3}
\begin{figure}[h]
  \centering

  
    \begin{subfigure}[t]{0.24\textwidth}
      \centering
      \includegraphics[width=\linewidth]{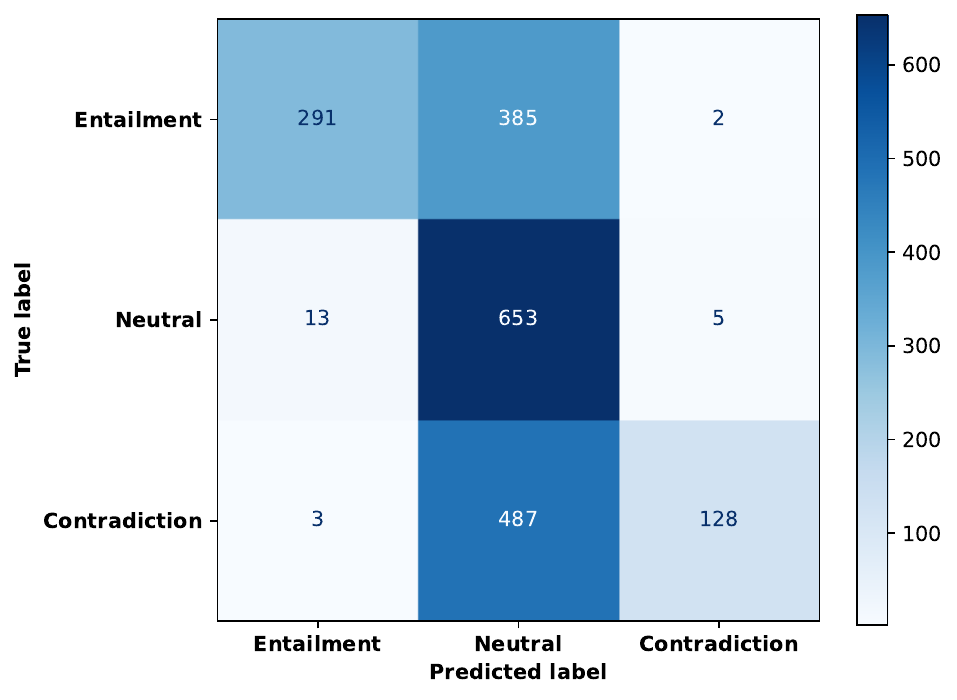}
      \caption{Qwen3-4B-Base}
    \end{subfigure}%
    \begin{subfigure}[t]{0.24\textwidth}
      \centering
      \includegraphics[width=\linewidth]{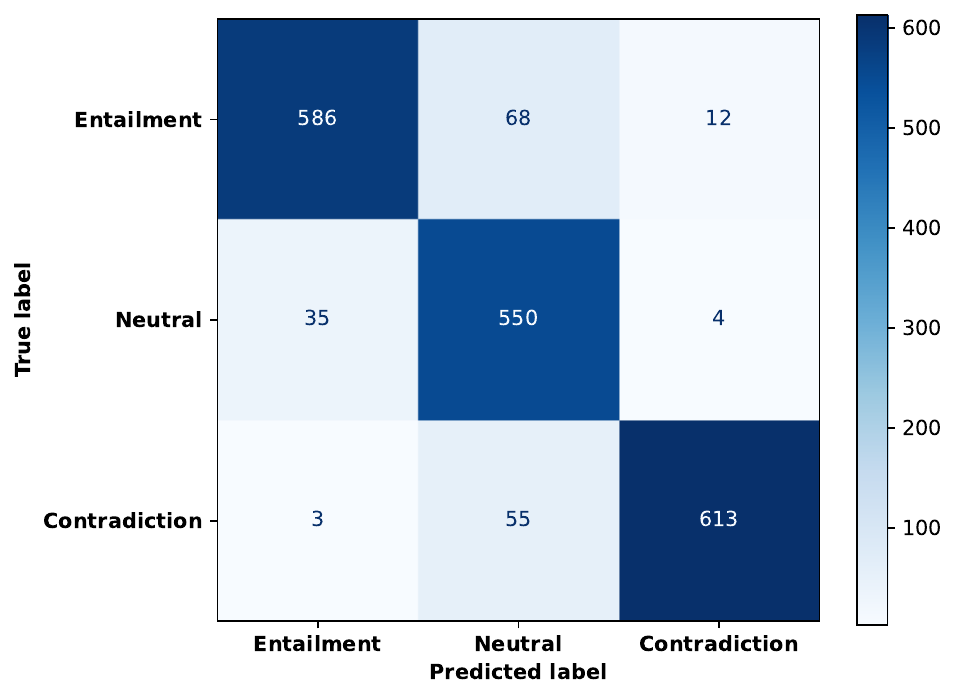}
      \caption{Qwen3-4B-Instruct}
    \end{subfigure}
    \caption{\textbf{The Confusion Matrix of NLI task on SNLI.}}
    \label{fig:confu}
  
\end{figure}


Since SSC-GRPO relies on the model’s self-consistency judgments to provide reward signals, it is important to evaluate its NLI capability. To complement the rollout-based evaluation in Section~\ref{sec:6.1}, we assess Qwen3-4B-Base and Qwen3-4B-Instruct on standard NLI benchmarks with ground-truth annotations, using the same NLI prompt employed for consistency evaluation.

We first evaluate the models on MathNLI~\citep{de2025math}. Table~\ref{tab:mathnli} reports accuracy, F1-score, and precision. Both models achieve high accuracy on this mathematical NLI task, with Qwen3-4B-Base reaching 83.30\% and Qwen3-4B-Instruct reaching 92.18\%. These results demonstrate their NLI classification capability in mathematical contexts and support using the models themselves for consistency judgment.

To further evaluate their NLI performance, we conduct a detailed analysis on SNLI~\citep{bowman-etal-2015-large}. Key metrics are reported in Table~\ref{tab:nli}, and the confusion matrices are presented in Figure~\ref{fig:confu}. The instruct model achieves high overall accuracy. For the base model, most errors involve classifying entailment or contradiction examples as neutral, while its entailment and contradiction predictions remain highly precise. In SSC-GRPO, neutral judgments receive a consistency score of zero, rather than an opposing entailment or contradiction score. Thus, although the base model misses some informative cases, the high precision of its non-neutral predictions supports their use as consistency signals. Together, these results provide further evidence of the models’ ability to perform NLI judgments independently.

\section{Case Study for SSC-GRPO}
To more clearly demonstrate that our method effectively mitigates context-sensitive factual hallucinations, we sample several cases from Qwen3-4B-base, as shown in Table~\ref{tab:case}. The original model makes errors even on simple addition and subtraction problems. However, after training with SSC-GRPO, the model successfully resolves these issues and no longer exhibits such errors.

\begin{table*}[h]
    \centering

         \resizebox{\linewidth}{!}{
        \begin{tabular}{p{24cm}}
    \toprule[1.2pt]
     \textbf{Question:} The proper divisors of $12$ are $1$, $2$, $3$, $4$ and $6$. A proper divisor of an integer $N$ is a positive divisor of $N$ that is less than $N$. 
     What is the sum of the proper divisors of the sum of the proper divisors of $284$? \\

     \textbf{Golden Answer} $284$
        \\
     \midrule
     
     \textbf{Vanilla:} \textcolor{red}{(Wrong)}

     First, I need to find the sum of the proper divisors of $284$. Then, I will find the proper divisors of that sum and calculate their sum.  find the sum of the proper divisors of $284$, I'll start by listing its proper divisors: $1$, $2$, $4$, $71$, and $142$. 
     The sum of these is $1 + 2 + 4 + 71 + 142 = 220$. Now I need to find the proper divisors of $220$ and then sum them up.\\
     
     The proper divisors of $220$ are $1$, $2$, $4$, $5$, $10$, $11$, $20$, $22$, $44$, $55$, and $110$. \\
     \colorbox{red!15}{The sum of these proper divisors is $1 + 2 + 4 + 5 + 10 + 11 + 20 + 22 + 44 + 55 + 110 = 304$}.
     
     So, the sum of the proper divisors of the sum of the proper divisors of $284$ is $304$.\\

     \colorbox{red!15}{\textless{}answer\textgreater{} 304 \textless{}/answer\textgreater{}}
    \\
     \midrule

     \textbf{SSC-GRPO:}\textcolor{green}{(Correct)}
     
     First, let\'s find the sum of the proper divisors of $284$. 
     
     To do that, we need to find all proper divisors of $284$, add them up, and then find the sum of the proper divisors of that sum.\\
     
     1. Find the proper divisors of $284$.  To find the proper divisors of $284$, we first factorize $284$. 
     \\
     $(284 = 2^2 \times 71)$.
     \\
     The divisors of $284$ are: $1$, $2$, $4$, $71$, $142$, $284$.  The proper divisors of 284 are all divisors except $284$ itself, so they are: $1$, $2$, $4$, $71$, $142$.\\
     
     2. Calculate the sum of the proper divisors of 284.$[   1 + 2 + 4 + 71 + 142 = 220   ]$
     3. Now, we need to find the sum of the proper divisors of $220$. First, find the proper divisors of $220$.
     To find the proper divisors of 220, we first factorize 220.   $(220 = 2^2 \times 5 \times 11)$
     The divisors of $220$ are: $1$, $2$, $4$, $5$, $10$, $11$, $20$, $22$, $44$, $55$, $110$, $220$. 
     The proper divisors of $220$ are all divisors except $220$ itself, so they are: $1$, $2$, $4$, $5$, $10$, $11$, $20$, $22$, $44$, $55$, , $110$.
     \\
     \colorbox{green!15}{4. Calculate the sum of the proper divisors of $220$.}\\
     \colorbox{green!15}{$[  1 + 2 + 4 + 5 + 10 + 11 + 20 + 22 + 44 + 55 + 110 = 284]$ }\\
     
     The sum of the proper divisors of the sum of the proper divisors of $284$ is $284$. \\
     
     \colorbox{green!15}{\textless{}answer\textgreater{} 284\textless{}/answer\textgreater{}}
     \\
     \bottomrule[1.2pt]
\end{tabular}} %
    \caption{\textbf{The case of our method in Qwen3-4B-base}}
    
    \label{tab:case}
\end{table*}

\section{Prompt Templates}
In this section, we provide detailed prompt templates used in various stages of our experiments.
These templates ensure the consistency and reproducibility of our experimental setup.

Figures~\ref{fig:sys} and~\ref{fig:user} show the system prompt and user prompt used in Section~\ref{sec:3.3} to detect the locations and types of hallucinations occurring during the model’s reasoning process. We select a strong reference model (Qwen3-235B-A22B in this paper) as the hallucination detector and use these prompts to perform detection, as well as to rewrite the hallucinated content into a standalone question when needed. The detector returns the hallucination position, its type, and the rewritten question derived from the hallucination (if applicable).

Figure~\ref{fig:prompt_gsm8k} presents the prompt used in Section~\ref{sec:3.4} for benchmark data augmentation with a strong reference model (also Qwen3-235B-A22B). We instruct the model to add additional conditions that commonly appear in reasoning contexts but do not affect the final result, and to return the augmented question along with its answer.

Figure~\ref{fig:prompt_nli} illustrates the prompt used during SSC-GRPO training for the policy model to perform self-consistency judgment. We formulate this judgment as an NLI task, provide few-shot examples, and require the model to output only the classification result.

\label{apdx:4}

\begin{figure*}[t]
  \centering
  \begin{promptbox}{System prompt for detecting hallucinations}

\textbf{Instruction: } 

You are a strict auditor of mathematical solutions and an expert in “hallucination localization/rewriting.” You will receive three segments: `\textless{}Question\textgreater{}', `\textless{}Standard\_Solutions\textgreater{}', and `\textless{}Model\_Solution\textgreater{}'. Output ONLY JSON as required.
\\ \\
\textless{}Input Data\textgreater{}

\textless{}Question\textgreater{}

\{question\}

\textless{}/Question\textgreater{}

\textless{}Standard\_Solutions\textgreater{}

\{solutions\}

\textless{}/Standard\_Solutions\textgreater{}

\textless{}Model\_Solution\textgreater{}

{model\_solution}

\textless{}/Model\_Solution\textgreater{}
\\ \\
\textless{}Task\textgreater{}

1) Determine whether the final answer is correct  
- Extract the standard final answer from `\textless{}Standard\_Solutions\textgreater{}' (prefer `\textbackslash boxed\{...\}'; otherwise take the last explicitly stated final value).  

- Extract the model’s final submitted answer from `\textless{}Model\_Solution\textgreater{}' (use the final conclusion/the last `\textbackslash boxed\{...\}'/“Final Answer ...” as the reference; ignore intermediate work). 

- Judge whether the two are mathematically equivalent. 

- If correct: output `is\_correct=true', and set all other fields to `null'.

2) If the final answer is incorrect: locate the “earliest hallucinated sentence that directly affects subsequent reasoning”  

- Treat `\textless{}Model\_Solution\textgreater{}' as consisting of multiple sentences (or multiple lines/steps). Audit it sentence by sentence.  

- Find the first sentence that will directly cause the subsequent reasoning to derail or the final answer to be wrong.  

- Return only this one sentence as `hallucinated\_sentence' (do not return multiple sentences).

3) Determine the hallucination type: context inconsistency vs factual error  

- Context inconsistency (`context inconsistency'): the sentence contradicts the problem conditions, previously established conclusions in the preceding derivation, symbol definitions, variable meanings, etc. (e.g., treating $x$ defined earlier as $y$; misstating a given condition; saying “$AB=5$” when “$AB=3$” is given).  

  - If it belongs to this type: set `is\_context\_inconsistency=true' and provide a brief `explanation' (indicate what contextual statement it conflicts with).  
  
- Factual error (`factual error'): the sentence itself contains a mathematically verifiable mistake in arithmetic/algebra/theorem usage, etc. (e.g., $7*8=54$; $(a+b)^2=a^2+b^2$; wrong differentiation/integration formula; illegal equation-solving step), and it does not primarily rely on a conflict with earlier definitions.  

  - If it belongs to this type: set `is\_context\_inconsistency=false' and continue to Step 4.

4) If it is a factual error: rewrite the hallucinated sentence into a new, “independently solvable” problem `hallucination\_question' 

- Goal: rewrite the erroneous step that caused the mistake into a standalone mini-problem, so that one can verify/derive the correct conclusion based on that mini-problem alone.

- The new problem must:  

  a) Be self-contained: include all required given conditions/variable definitions/expressions, without relying on the original problem or other parts of the original reasoning;  
  
  b) Be focused: test only the key fact/calculation/theorem involved in that sentence;  
  
  c) Be decidable: lead to a clear numeric or symbolic answer.  
  
- Output only the rewritten “problem text,” with no answer and no solution.

One-shot examples (for understanding how to “rewrite into an independent problem”; do not copy)  

[Hallucinated sentence] “Given $f(x)=x^3-3x$, we have $f'(x)=3x^3-3$.”  

[Rewritten standalone problem] “Let $f(x)=x^3-3x$. Find and simplify the expression for the derivative $f'(x)$.”

Another example closer to an algebraic chain:  

[Hallucinated sentence] “If $a+b=5$ and $ab=6$, then $a^2+b^2=11$.”  

[Rewritten standalone problem] “Given real numbers a,b such that $a+b=5$ and $ab=6$, find the value of $a^2+b^2$.”
\\ \\
\textless{}Output Requirements\textgreater{} 

- If `is\_correct=true': all other fields must be `null'.

- If `is\_correct=false': you must provide `hallucinated\_sentence', `is\_context\_inconsistency', and `explanation';  

  - If `is\_context\_inconsistency=false' (factual error): you must also provide `hallucination\_question';  
  
  - If `is\_context\_inconsistency=true': `hallucination\_question' must be `null'.  
  
- Output JSON only; no extra text is allowed.

\textless{}Output JSON Schema\textgreater{}

\{

~~~~  “is\_correct": boolean,
  
~~~~  “standard\_answer": string or null,
  
~~~~  “model\_answer": string or null,

~~~~  “hallucinated\_sentence": string or null,
  
~~~~  “is\_context\_inconsistency": boolean or null,
  
~~~~  “explanation": string or null,
  
~~~~  “hallucination\_question": string or null
  
\}

  \end{promptbox}
  \caption{System prompt for detecting hallucinations during reasoning in Section~\ref{sec:3.3}}
  \label{fig:sys}
\end{figure*}

\begin{figure*}[t]
  \centering
  \begin{promptbox}{User Prompt for detecting hallucinations}

\textbf{Input}:

Please audit the following problem, standard solution, and model solution according to the system instructions, and output JSON only.
\\ \\
\textless{}Question\textgreater{}

\{question\}

\textless{}/Question\textgreater{}

\textless{}Standard\_Solutions\textgreater{}

\{solutions\}

\textless{}/Standard\_Solutions\textgreater{}

\textless{}Model\_Solution\textgreater{}

{model\_solution}

\textless{}/Model\_Solution\textgreater{}
  \end{promptbox}
  \caption{User prompt for detecting hallucinations during reasoning in Section~\ref{sec:3.3}}
  \label{fig:user}
\end{figure*}

\begin{figure*}[t]
  \centering
  \begin{promptbox}{Context-Augmented data Prompt}
\textbf{Instruction:} 

You will receive a JSON corresponding to a single “question–answer” sample. Your task is, **without changing the correctness of the answer**, to perform “same-domain, misleading distractor-style augmentation” **only on** the `question`, to simulate that this question is **one sub-step** within a larger problem-solving process; the newly added content represents information/intermediate conclusions from “other parts of the full problem” but **is irrelevant to the current question being solved**.
 \\
 \\
**[Input Format]**

- A JSON containing at least:

  - ``question": \textless{}original question text\textgreater{}
  
  - ``answer": \textless{}original answer text/value/expression\textgreater{}
  
- It may also include other fields. **Do not assume field names**, and **do not add or remove any fields**.
\\
\\
**[Augmentation Goals]**

In `question':

1) Add several conditions/background/intermediate conclusions in the **same domain** as the original problem (from “other parts”), which should be **misleading** but **irrelevant to solving the current question**, and must not change the original answer.

2) Create the sense that the “sub-question is embedded in a larger context,” but **do not use** explicit phrases such as “big problem/subproblem/step”; embed only via natural sentences.

3) **Keep only one question**, and the **last sentence** must ask that question; **keep the asking style consistent with the original** (e.g., if the original is “Compute…/What is…,” the augmented version must follow the same paradigm).

4) Do not introduce conflicting conditions that would change the answer: do not alter given dimensions/definitions/constraints; do not replace key numeric values; you may introduce “seemingly related but actually irrelevant” definition reminders, side theorems, neighboring concepts, approximate-but-useless parameters, historical statistics, boundary-case discussions, etc.

5) Add 1–3 distractors of moderate length; preferably insert them in the front or middle of the question body with natural transitions.
\\
\\
**[Invariance \& Consistency]**

- The `answer' **must match the input exactly** (character-for-character, including formatting/units/case).

- If after augmentation you find the answer might change or the reasoning is affected, **rewrite the distractor content** until it no longer affects the answer.

- Ensure there is **only one question mark** (Chinese “?” or English “?”) in the `question', and it appears in the last sentence; avoid creating new independent interrogative sentences.
\\
\\
**[Language \& Format]**

- **Follow the input language**: if the original `question` is in English, the augmented `question` must be in English; otherwise use Chinese.

- **Keep the JSON structure, key names, and field order exactly the same as the input**; modify only the `question` field; copy all other fields unchanged.

- **Output strictly JSON only**—no explanations, prefixes/suffixes, code fences, or extra text.
\\
\\
**[Output Requirements]**

- Output a JSON for the augmented sample:

  - `question': the revised question with inserted distractors satisfying all requirements.
  
  - `answer': exactly the same as the input (must not be changed).
  
  - Any other fields: identical to the input (must not be changed).
\\ \\

\textbf{Input}:

\{

~~~~``question": \{question\},
    
~~~~``answer": \{answer\}

\}    
  \end{promptbox}
  \caption{Prompt for augmented dataset in section~\ref{sec:3.4}}
  \label{fig:prompt_gsm8k}
\end{figure*}

\begin{figure*}[t]
  \centering
  \begin{promptbox}{NLI Judge Prompt}
\textbf{Instruction:} 

You are performing a Natural Language Inference (NLI) task.
Given a dictionary containing a ``premise" and a ``hypothesis", determine the relationship between them.
You must answer with ONLY one word, chosen from exactly these three labels:

entailment, contradiction, neutral
\\ \\
No explanation.

No extra words.

Only output the label.
\\
Below are examples:

Example 1:

\{``premise": ``A dog is running through a field.",

 ``hypothesis": ``An animal is outdoors."\}
 
Answer: entailment
\\ \\
Example 2:

\{``premise": ``A woman is cooking in the kitchen.",

 ``hypothesis": ``No one is inside the house."\}
 
Answer: contradiction
\\ \\
Example 3:
\{``premise": ``Two people are talking in the park.",

 ``hypothesis": ``They are discussing politics."\}
 
Answer: neutral
\\ \\
Now classify the following sample strictly in the same format:

\textbf{Input:}

\{

~~~~``premise": \textless{}premise\textgreater{},

~~~~ ``hypothesis": \textless{}hypothesis\textgreater{}
 
 \}
 
\end{promptbox}
  \caption{Prompt for NLI judge during training}
  \label{fig:prompt_nli}
\end{figure*}

\end{document}